\RequirePackage[T1]{fontenc}
\documentclass[letterpaper,10pt,conference]{ieeeconf}
\IEEEoverridecommandlockouts
\usepackage{mathptmx,amsmath,amssymb}
\usepackage{graphicx,booktabs,cite,placeins,flushend}
\usepackage[hidelinks]{hyperref}
\newcommand{\SE}{\mathrm{SE}(3)}
\newcommand{\R}{\mathbb{R}}
\title{\LARGE\bf ReCo: Response-Consistent Locomotion with Policy-Aware MPC for Legged Manipulation}
\author{Kuankuan Sima$^{1}$, Yichao Gao$^{1}$, Chenxi Gu$^{2}$, Kefan Zhao$^{2}$, and Lin Zhao$^{1}$\\National University of Singapore%
\thanks{$^{1}$Kuankuan Sima, Yichao Gao, and Lin Zhao are with the Department of Electrical and Computer Engineering, National University of Singapore, Singapore 119077, Singapore (e-mail: {\tt\footnotesize kuankuan\_sima@u.nus.edu; elezhli@nus.edu.sg}).}%
\thanks{$^{2}$Chenxi Gu and Kefan Zhao are with the Department of Mechanical Engineering, National University of Singapore, Singapore 119077, Singapore.}%
}
\begin{document}
\maketitle
\thispagestyle{empty}
\pagestyle{empty}
\suppressfloats[t]
\begin{abstract}
Continuous legged manipulation requires accurate end-effector tracking while the base keeps walking. Combining reinforcement learning (RL) with model predictive control (MPC) suits this task: the learned policy provides robust locomotion, while MPC coordinates the base and arm to compensate for tracking errors. However, MPC can compensate only for base motion that it can predict, and a learned policy's command response varies with gait phase, contact, and payload. We present ReCo, a framework that couples \underline{re}sponse-\underline{co}nsistent locomotion with policy-aware MPC for legged manipulation. Response shaping trains the policy to respond to commands consistently and repeatably across randomized dynamics. An identified closed-loop response model then lets MPC jointly plan locomotion commands and arm motion. On the simulation benchmark, ReCo reduces position and orientation root-mean-square error (RMSE) by 28.7\% and 27.4\% relative to the best baseline for each metric. Real-world experiments demonstrate onboard continuous legged manipulation with coordinated base and arm motion.
\end{abstract}
\section{Introduction}
\label{sec:intro}
\begin{figure}[t]
\centering
\includegraphics[width=\columnwidth]{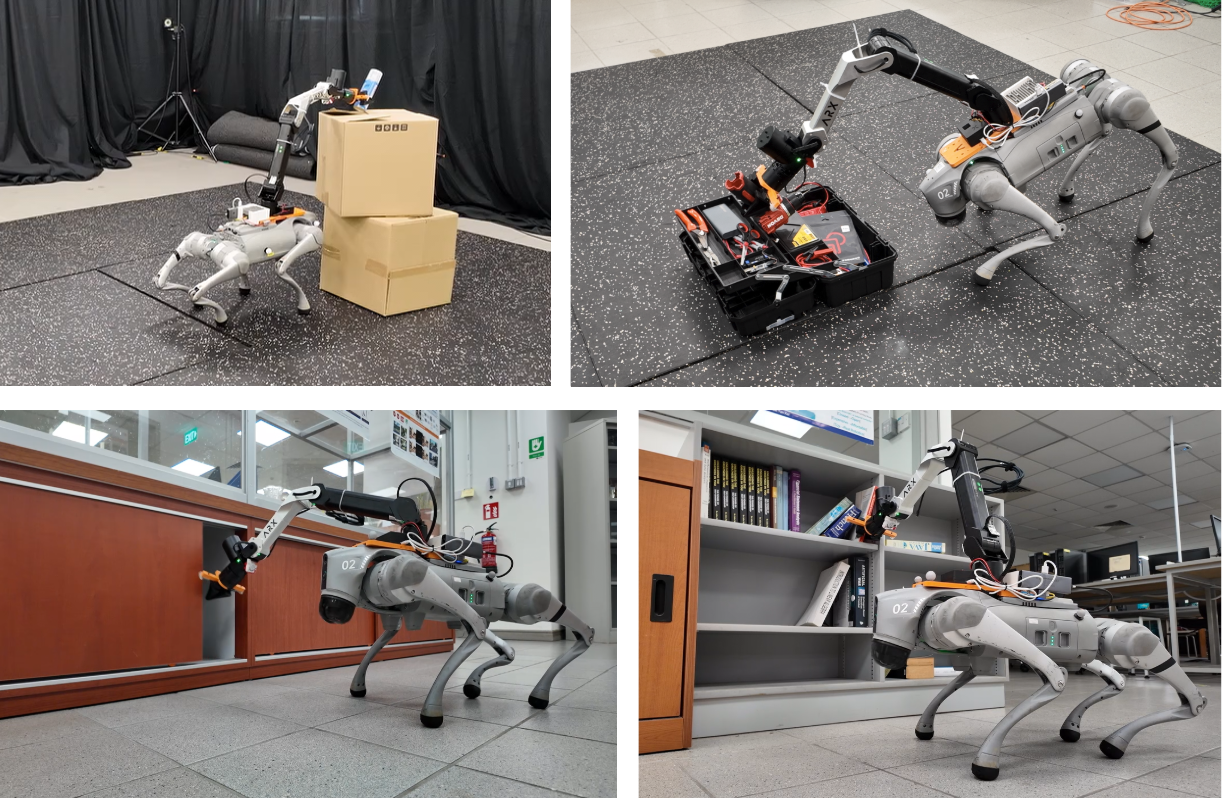}
\caption{Demonstration of continuous legged manipulation tasks in the real world. ReCo jointly plans locomotion commands and arm motions to track task-space end-effector trajectories for object organization, cabinet-door opening, and pick-and-place.}
\label{fig:teaser}
\end{figure}

A quadruped equipped with an arm can bring manipulation capabilities to environments that are difficult for conventional mobile platforms to traverse. Tasks such as inspection, scanning, and tool positioning can require the robot to track a world-frame end-effector (EE) trajectory while walking. Every step moves the arm's supporting frame: base translation, attitude changes, and gait oscillations all propagate to the EE. Continuous legged manipulation therefore requires coordination throughout locomotion, beyond repositioning the base to extend the arm's workspace (Fig.~\ref{fig:teaser}).

Robust locomotion and precise manipulation place different demands on control. Reinforcement learning (RL) has enabled quadrupeds to traverse challenging terrain under uncertain contact conditions and dynamics~\cite{hwangbo2019agile,lee2020terrain,miki2022perceptive,margolis2023wtw}. Model-based control explicitly represents task geometry, motion constraints, and future references, supporting coordinated locomotion and manipulation~\cite{bellicoso2019alma,pankert2020perceptive}. Learned whole-body controllers also demonstrate effective leg--arm coordination~\cite{fu2023deep,pan2024roboduet}. However, (i) model-based control relies on accurate contact and dynamics models, which are hard to maintain on uneven terrain with uncertain contact, and (ii) learned whole-body control balances task-space accuracy and stability within one reward, which can require retuning across terrains, payloads, and disturbances.

A modular alternative assigns locomotion to RL and base--arm coordination to model predictive control (MPC)~\cite{ma2022combining}. In such architectures the key difficulty lies at the interface: the manipulation planner must predict how the policy-controlled base will move under candidate locomotion commands. To MPC, the base is a closed-loop subsystem whose command response can vary with gait phase, contact conditions, and payload. To the locomotion policy, arm motion changes the dynamics it must stabilize. Assuming that a commanded velocity or posture is realized immediately neglects these interactions. Even an accurate mean-velocity model can miss the periodic height and attitude motion that propagates to the EE.

Our key insight is that \emph{a locomotion policy should be trained not only to be robust but also to be predictable by the planner that commands it}. Locomotion supplies sustained displacement and disturbance recovery, while the arm supplies fast local corrections within its workspace and actuator limits; a planner can exploit this division of labor only if the base response is repeatable. A consistent delay can be compensated by feedforward, whereas a variable delay cannot. We therefore seek a locomotion response that a low-order model can predict over a useful horizon.

We present ReCo, a framework that couples response-consistent locomotion with policy-aware MPC for legged manipulation. Our contributions are:
\begin{itemize}
\item A training method that makes the command response of a locomotion policy repeatable through reference tracking, phase repeatability, steady-state gain, and cross-domain consistency rewards.
\item A policy-aware MPC that jointly plans locomotion commands and arm motion for continuous world-frame EE $\SE$ tracking, where an identified closed-loop response model with a gait-periodic component serves as the predictive interface to the learned policy.
\item Simulation and hardware experiments in which ReCo reduces position and orientation root-mean-square error (RMSE) by 28.7\% and 27.4\% relative to the best of five baselines for each metric, response shaping lowers normalized prediction mean squared error (MSE) by 59.6\%, and the full system runs onboard.
\end{itemize}

\section{Related Work}
\label{sec:related}
\subsection{Model-Based Legged Manipulation}
MPC enables legged locomotion through ground-reaction-force optimization~\cite{dicarlo2018cheetah} and terrain-aware motion planning~\cite{grandia2022perceptive}. For legged manipulation, ALMA combines online planning, torso posture optimization, and hierarchical whole-body control~\cite{bellicoso2019alma}; perceptive MPC incorporates environmental geometry into continuous mobile manipulation~\cite{pankert2020perceptive}. These approaches coordinate base and arm motion through explicit models and constraints. When a learned policy executes locomotion, the manipulation planner additionally needs to predict the command-to-motion response of the policy--robot closed loop; ReCo supplies this prediction with an identified closed-loop response model.

\subsection{Learning-Based Locomotion and Manipulation}
Actuator modeling, domain randomization, and disturbance training support robust sim-to-real locomotion~\cite{tan2018sim,hwangbo2019agile}. Proprioceptive and perceptive policies extend locomotion to challenging terrain~\cite{lee2020terrain,miki2022perceptive}, while online adaptation~\cite{kumar2021rma} and parallel training~\cite{rudin2022minutes} improve robustness and training efficiency. Walk These Ways provides a single policy with gait, posture, and velocity commands~\cite{margolis2023wtw}. ReCo adopts this command interface and shapes the policy's transient and periodic responses for predictive planning.

Deep Whole-Body Control learns a unified leg--arm policy~\cite{fu2023deep}, and RoboDuet learns cooperating locomotion and manipulation policies~\cite{pan2024roboduet}. Visual whole-body control combines a visuomotor policy with a controller conditioned on base velocity and EE pose~\cite{liu2024visual}. UMI on Legs learns to track task-frame manipulation trajectories with future-target observations~\cite{ha2024umilegs}. Portela et al. demonstrate EE pose tracking over a large workspace and rough terrain~\cite{portela2025pose}, and learned force control supports compliant interaction~\cite{portela2024force}. ReCo focuses on the predictive interface between a learned locomotion policy and policy-aware MPC for joint optimization of locomotion commands and arm motion.

\subsection{Hybrid Control and Policy Response Modeling}
Ma et al. condition a learned locomotion policy on predicted external wrenches from model-based manipulation~\cite{ma2022combining}. Their interface lets locomotion anticipate arm-induced disturbances; ReCo lets policy-aware MPC anticipate the base response to candidate commands. RAMBO learns feedback corrections to model-based whole-body control~\cite{cheng2025rambo}; in ReCo, the learned component is the locomotion policy, and the response model is fitted to it.

Learning-based MPC can improve dynamics models, tune control objectives, or impose predictive constraints on learned behavior~\cite{hewing2020learning}. Kim et al. identify lifted linear dynamics of an RL-controlled biped for MPC-based navigation~\cite{kim2026koopman}. ReCo extends policy-response modeling by training the policy to produce responses that a low-order model can predict and by representing the gait-periodic height and attitude variations that affect manipulation.

\section{Method}
\label{sec:method}
\begin{figure}[t]
\centering
\includegraphics[width=\columnwidth]{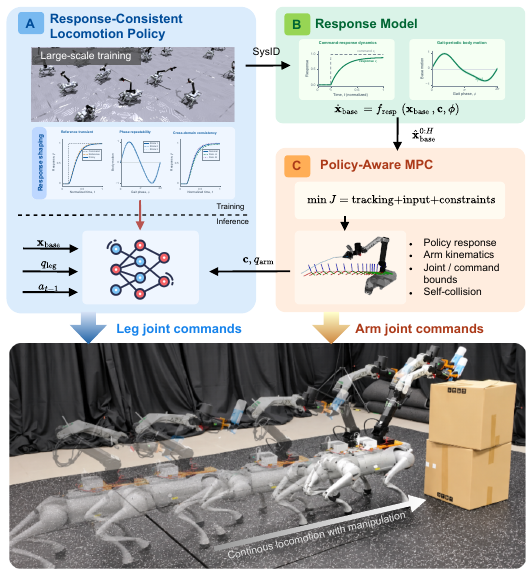}
\caption{Overview of ReCo. Response shaping and model identification produce a closed-loop response model of the locomotion policy. MPC integrates this model, arm kinematics, and state feedback to jointly plan locomotion commands and arm motion.}
\label{fig:architecture}
\end{figure}

ReCo trains a locomotion policy with a consistent command response, identifies a closed-loop response model of that policy, and uses the model in a policy-aware MPC that jointly plans locomotion commands and arm motion (Fig.~\ref{fig:architecture}).

\subsection{Problem Formulation}
We track a time-indexed world-frame EE pose reference $T_{WE}^{\mathrm{ref}}(t)\in\SE$ during locomotion. Frames $W$, $B$, and $E$ denote the world, base, and EE; $T_{AB}$ maps coordinates from frame $B$ to frame $A$. Let $q\in\R^6$ denote arm joint positions. MPC selects locomotion commands and arm joint velocities.

A gait-conditioned policy controls the legs using proprioception, commands, and gait timing. Its command vector is $\mathbf c=(c_x,c_y,c_\omega,c_h,c_\theta)\in\R^5$, comprising planar velocity, yaw rate, height, and pitch. The base prediction state is
\begin{equation}
 \mathbf x_{\mathrm{base}}=(p_W,\psi,\theta,\rho,v_x,v_y,\omega_z)\in\R^9,
 \label{eq:base-state}
\end{equation}
where $p_W\in\R^3$ is the modeled base position in $W$, $(\psi,\theta,\rho)$ are ZYX yaw, pitch, and roll, $(v_x,v_y)$ are heading-frame velocities, and $\omega_z$ is body-frame angular velocity about the $z$ axis. The policy has no roll command; roll remains a predicted state because it affects the EE pose. With gait phase $\phi$, the full state and input are
\begin{equation}
 x=(\mathbf x_{\mathrm{base}},\phi,q)\in\R^{16},\qquad
 u=(\mathbf c,u_q)\in\R^{11},
 \label{eq:state-input}
\end{equation}
where $u_q\in\R^6$ denotes arm joint velocity commands. The response model in Sec.~\ref{sec:response-model} predicts the motion produced by the policy--robot closed loop under candidate commands.

\subsection{Response-Consistent Locomotion Training}
We train the locomotion policy with proximal policy optimization~\cite{schulman2017ppo} and the gait-conditioned interface of Walk These Ways~\cite{margolis2023wtw}. The gait setting is fixed over each prediction horizon. Response shaping denotes the training terms used to encourage response consistency: repeatable command transients and gait-phase-dependent body motion across randomized dynamics. These terms act on $y=(v_x,v_y,\omega_z,h,\theta)$, where $h$ is body height measured relative to its nominal value. Each of these five command channels drives a critically damped reference generator,
\begin{equation}
 \ddot\xi_i=\lambda_i^2(c_i-\xi_i)-2\lambda_i\dot\xi_i,
 \qquad |\dot\xi_i|\leq r_i,
 \label{eq:reference}
\end{equation}
where $\xi_i$ is the target response and $\lambda_i$ and $r_i$ specify its bandwidth and rate limit. Unsaturated dynamics use an exact zero-order-hold update; saturation produces a rate-limited ramp. The height command $c_h$ uses the same nominal offset as $h$.

To separate command transients from gait oscillations, we use $\tilde y=y-\hat\delta(\phi,\nu)$, where $\nu$ is planar speed and $\hat\delta$ estimates the gait-periodic component of $y$ during training. This estimate uses running means over phase and speed bins, updated at each environment step from nominal-domain environments. The reference-tracking reward is
\begin{equation}
 r_{\mathrm{ref}}=g\,
 \frac{\sum_i w_i\exp[-(\tilde y_i-\xi_i)^2/\sigma_i^2]}{\sum_i w_i},
 \label{eq:response_reward}
\end{equation}
where $w_i$ and $\sigma_i$ are channel weights and error scales. The gate $g\in\{0,1\}$ disables this reward during external disturbances and a post-disturbance interval, allowing recovery to take precedence over transient matching.

\begin{table}[t]
\centering
\caption{Response-shaping terms. The reference reward contributes to $r_+$; the three nonnegative losses enter $r_-$ with negative weights.}
\label{tab:rewards}
\small
\setlength{\tabcolsep}{3pt}
\renewcommand{\arraystretch}{1.15}
\begin{tabular}{@{}p{0.29\columnwidth}p{0.66\columnwidth}@{}}
\toprule
Term & Definition and purpose\\
\midrule
Reference tracking & $r_{\mathrm{ref}}$ in~\eqref{eq:response_reward}; matches the commanded transient.\\
Phase repeatability & $\ell_{\mathrm{phase}}=\sum_i \mu_i^{\mathrm{ph}}w_i^{\mathrm{ph}}(y_i-\bar y_i-\hat\delta_i)^2$; regularizes the gait waveform.\\
Steady-state gain & $\ell_{\mathrm{gain}}=\sum_i \mu_i^{\mathrm{ss}}w_i^{\mathrm{ss}}(\tilde y_i-c_i)^2$; discourages settled command bias.\\
Cross-domain consistency & $\ell_{\mathrm{domain}}^d$ in~\eqref{eq:twin}; matches randomized instances to the nominal response.\\
\bottomrule
\end{tabular}
\par\smallskip
\begin{minipage}{\columnwidth}\footnotesize
$\bar y$ is the low-pass local mean. Channel masks $\mu_i^{\mathrm{ph}}$ and $\mu_i^{\mathrm{ss}}$ activate after settling; $w_i^{\mathrm{ph}}$ and $w_i^{\mathrm{ss}}$ weight channels. These terms supplement the underlying locomotion rewards and regularization.
\end{minipage}
\end{table}

Table~\ref{tab:rewards} summarizes the response-shaping terms. Phase regularization penalizes deviations from the repeatable waveform, rather than its amplitude. A separate steady-state term discourages command bias. For cross-domain consistency, groups of environments receive identical command sequences and gait phases. Each group contains a nominal environment and instances with randomized mass, payload, friction, and actuator strength. Instance $d$ incurs
\begin{equation}
 \ell_{\mathrm{domain}}^d=m^d\|\tilde y^d-\tilde y^{\mathrm{nom}}\|_{W_d}^2,
 \label{eq:twin}
\end{equation}
where $\|a\|_W^2=a^\top Wa$ and $W_d$ weights all five response channels, including height. The mask $m^d$ excludes nominal, invalid, or desynchronized comparisons. The nominal environment supplies the reference response for the randomized instances, while~\eqref{eq:response_reward} specifies the target transient.

These terms augment the locomotion reward. Positive terms $r_+\geq0$ and signed penalties $r_-\leq0$ are combined as $r=r_+\exp(c_-r_-)$, with $c_->0$~\cite{margolis2023wtw}. Training first establishes locomotion, then introduces reference tracking, consistency penalties, broader randomization, and stronger disturbances.

\subsection{Closed-Loop Response Model}
\label{sec:response-model}
The response model combines command-response dynamics with gait-periodic base motion. Velocity channels follow
\begin{equation}
 \dot v_i=(k_i c_i-v_i)/\tau_i,
 \qquad i\in\{x,y,\omega\},
 \label{eq:velocity}
\end{equation}
where $v_\omega=\omega_z$, $k_i$ is the response gain, and $\tau_i$ is the time constant. The second-order training reference in~\eqref{eq:reference} specifies a target transient; it does not impose the order of the realized closed loop. We use one response state per velocity channel to keep horizon optimization compact.

For $\eta=(z,\theta,\rho)$, where $z$ is the world height of the modeled base frame at the arm mounting plane, we add a speed-dependent gait-periodic component, modeled as a harmonic series:
\begin{equation}
\begin{aligned}
 \delta_j&=\sum_{n=1}^{N_h}(a_{0jn}+a_{1jn}\nu)\sin(n\phi+\beta_{jn}),\\
 \dot\eta_j&=\frac{b_j+k_jc_j-(\eta_j-\delta_j)}{\tau_j}
               +\dot\delta_j,\qquad \dot\phi=2\pi f_g,
\end{aligned}
\label{eq:gait}
\end{equation}
where $j\in\{z,\theta,\rho\}$, $c_z=c_h$, $f_g$ is gait frequency, and the roll channel has no command input ($k_\rho=0$). The smoothed planar speed is $\nu=\sqrt{v_x^2+v_y^2+\epsilon}$ with a small $\epsilon>0$. The coefficients $a_{0jn}$ and $a_{1jn}$ set the fixed and speed-dependent amplitudes, and $\beta_{jn}$ sets the phase offset. We use $N_h=2$ for stride and half-stride components. Only $b_z$ is nonzero: it sets the nominal mounting-plane height for the policy's zero height command, $c_h=0$.

The gait-periodic component $\delta$ is identified offline for prediction, separately from the training-time estimate $\hat\delta$. The state $\eta$ includes gait oscillations, so the lag acts on $\eta-\delta$, and $\dot\delta$ accounts for phase and speed variation. Planar world-frame velocity is obtained by yaw-rotating $(v_x,v_y)$, and the ZYX yaw rate satisfies
\begin{equation}
 \dot\psi=\frac{\omega_z+\sin\rho\,\dot\theta}{\cos\theta\cos\rho}.
 \label{eq:yaw-rate}
\end{equation}
The model captures the empirical closed-loop response without explicitly modeling arm-induced wrenches.

Parameters are identified offline for each frozen policy by bounded nonlinear least squares on multi-step open-loop prediction errors. Step, ramp, and chirp commands excite the response across randomized domains. Identification data are separate from the prediction and tracking runs in Sec.~\ref{sec:experiments}.

\subsection{Policy-Aware Model Predictive Control}
The response model and arm kinematics $\dot q=u_q$ define $\dot x=f(x,u)$. Forward kinematics gives $T_{WE}=T_{WB}T_{BE}(q)$, where $T_{BE}$ includes the arm mounting transform. Let $p_{WE}$ and $R_{WE}$ denote the position and rotation components of $T_{WE}$. At each update, MPC solves
\begin{equation}
\begin{aligned}
 \min_{x(\cdot),u(\cdot)}\quad &\Phi(x(t+H))+
     \int_t^{t+H}\!\left[\ell(x,u,t')+\mathcal P(x,u)\right]dt'\\
 \mathrm{s.t.}\quad &\dot x=f(x,u),\qquad x(t)=\hat x(t),
\end{aligned}
\label{eq:mpc}
\end{equation}
where $\hat x(t)$ is the current state estimate, $H$ is the horizon, and $\mathcal P$ contains soft penalties for joint bounds, command bounds, and selected self-collision distances. The stage cost is
\begin{equation}
\begin{split}
 \ell={}&\tfrac12\|e_p\|_{Q_p}^2+\tfrac12\|e_R\|_{Q_R}^2
       +\tfrac12\|\dot p_{WE}-\dot p_{WE}^{\mathrm{ref}}\|_{Q_v}^2\\
       &+\tfrac12\|u\|_{R_u}^2+\ell_{\mathrm{base}}+\ell_{\mathrm{posture}},
\end{split}
\label{eq:cost}
\end{equation}
where $Q_p,Q_R,Q_v$, and $R_u$ are cost-weight matrices and $e_p=p_{WE}-p_{WE}^{\mathrm{ref}}$. For the predicted and reference EE orientations, represented by unit quaternions $(w,\mathbf v)$ and $(w_r,\mathbf v_r)$,
\begin{equation}
 e_R=w\mathbf v_r-w_r\mathbf v+\mathbf v\times\mathbf v_r.
 \label{eq:orientation-error}
\end{equation}
The reference quaternion is sign-aligned with the prediction when $w_rw+\mathbf v_r^\top\mathbf v<0$. This quaternion residual is used for optimization; evaluation uses the geodesic rotation angle. Positions are linearly interpolated, orientations use quaternion interpolation, and reference velocities follow the interpolated position slopes. The terminal cost $\Phi$ penalizes EE pose error.

The quadratic term $\ell_{\mathrm{base}}$ penalizes pitch, roll, yaw rate, and command--response discrepancies; $\ell_{\mathrm{posture}}$ penalizes arm deviations from a nominal compact configuration. Base and arm decisions are coupled through the predicted EE pose, allowing the arm to anticipate command lag and gait oscillations.

We solve~\eqref{eq:mpc} with OCS2~\cite{farshidian2017ocs2} multiple-shooting sequential quadratic programming (SQP) and automatic differentiation, warm-starting each update from the previous solution. A reference manager samples the plan faster than the MPC update rate and holds locomotion commands between updates. Feedback latency is compensated by propagating the measured state over one sampling interval before optimization. The feedback comprises the base state, arm joint positions, and gait phase from the policy clock. Phase is wrapped to $[0,2\pi)$ in the shooting constraints.

\section{Experiments}
\label{sec:experiments}
We first compare ReCo with alternative control architectures, then isolate the effects of response shaping and model identification. Prediction and repeatability experiments examine how shaping affects base responses, a further experiment evaluates the gait-periodic component of the response model, and hardware demonstrations evaluate onboard execution.

\subsection{Experimental Setup}
\label{sec:setup}
\textit{Benchmark.} We evaluate a Unitree Go2 with a six-DoF X5 arm in MuJoCo~\cite{todorov2012mujoco} at a 2.5~ms integration step. The deterministic $16\times16$~m rolling heightfield has elevations between $-0.10$ and $0.10$~m and RMS height 0.0438~m. The benchmark contains 50 world-frame EE pose trajectories generated from randomly sketched paths with varied geometry and timing. Targets can lie outside the arm workspace at a fixed base pose, requiring locomotion to make them reachable. These trajectories test sustained base--arm coordination across diverse motion demands. Each controller executes every trajectory nominally and with an 80~N world-$+X$ force applied at the base center of mass for 0.1~s (8~N\,s impulse), yielding 100 runs per controller. References and their timing are fixed across controllers.

\textit{Baselines.} We compare representatives of three control paradigms. \emph{RL+MPC:} Ma et al.~\cite{ma2022combining} combine learned locomotion with model-based manipulation. We reimplement their approach with predicted wrench inputs and the corresponding locomotion training, using the same OCS2 infrastructure, horizon, and arm model as ReCo. This provides a closely related hybrid comparison with a different locomotion--manipulation interface. \emph{Model-based:} Pure MPC~\cite{qmcontrol} coordinates the whole body without a learned locomotion policy. \emph{Learned whole-body control:} RoboDuet~\cite{pan2024roboduet}, UMI on Legs~\cite{ha2024umilegs}, and Visual Whole-Body Control (Visual WBC)~\cite{liu2024visual} are retrained from their public implementations for Go2--X5, retaining their reward and observation designs and matching our training budget.

\textit{Command interfaces.} RoboDuet and Visual WBC require externally supplied base velocities; their evaluated low-level controllers do not generate these commands directly from a world-frame EE trajectory. We therefore pair both with a classical omnidirectional pure-pursuit (PP) tracker and convert EE targets to their native body-frame interfaces, denoting the results RoboDuet+PP and Visual WBC+PP. Table~\ref{tab:formal-benchmark} groups these two adaptations separately. UMI on Legs receives task-frame trajectory chunks and coordinates base and arm motion through its learned controller. All methods use the same reference clock; imposed trajectories replace high-level visual or task policies.

\textit{Training and metrics.} All learned controllers are trained in Isaac Gym~\cite{makoviychuk2021isaac} and evaluated in MuJoCo. The ReCo locomotion policy converges in approximately 10~h on one RTX~4090 with about 12~GB GPU memory. Position error is $\|p_{WE}-p_{WE}^{\mathrm{ref}}\|_2$; orientation error is the geodesic angle of $(R_{WE}^{\mathrm{ref}})^\top R_{WE}$. We report means and sample standard deviations of per-run RMSE over all 100 runs. For early failures, RMSE covers the recorded portion of the run and must be read together with falls and final progress. Success requires at least 97\% final progress without a fall, numerical fault, or early termination, endpoint position and orientation errors within 0.15~m and $40^\circ$ held continuously for at least 0.5~s, and at least 40\% of samples within both tolerances.

\textit{Implementation.} MPC runs at 100~Hz and uses a $1.0$~s horizon with $67$ shooting nodes at approximately $15$~ms spacing. Each update performs three SQP iterations with RK2 integration on three threads. Base-state feedback comes from the simulator in simulation and from the leg-kinematics odometry on hardware.

\subsection{Trajectory Tracking Performance}
\label{sec:benchmark}
\begin{table*}[t]
\centering
\caption{Uneven-terrain MuJoCo benchmark: 50 trajectories under nominal and push conditions (100 runs per controller). Errors are per-run RMSE means $\pm$ sample standard deviations, including recorded prefixes of early failures. PP denotes the external omnidirectional pure-pursuit tracker.}
\label{tab:formal-benchmark}
\small
\setlength{\tabcolsep}{5pt}
\begin{tabular}{llccccc}
\toprule
Controller & Paradigm & \shortstack{Position\\RMSE (m)$\downarrow$} & \shortstack{Orientation\\RMSE ($^\circ$)$\downarrow$} & \shortstack{Progress\\(\%)$\uparrow$} & Success$\uparrow$ & Falls$\downarrow$\\
\midrule
Ma et al.~\cite{ma2022combining} & RL+MPC & 0.199 $\pm$ 0.114 & 32.7 $\pm$ 14.6 & 82.6 & 31/100 & 8/100\\
Pure MPC~\cite{qmcontrol} & Model-based & 0.189 $\pm$ 0.038 & 46.8 $\pm$ 14.2 & 80.5 & 19/100 & 12/100\\
UMI on Legs~\cite{ha2024umilegs} & Learning & 0.883 $\pm$ 0.832 & 81.1 $\pm$ 34.3 & 12.9 & 12/100 & 80/100\\
\midrule
RoboDuet~\cite{pan2024roboduet}+PP & Learning & 0.679 $\pm$ 0.603 & 79.9 $\pm$ 29.0 & 65.7 & 42/100 & 9/100\\
Visual WBC~\cite{liu2024visual}+PP & Learning & 1.454 $\pm$ 1.101 & 68.2 $\pm$ 30.1 & 25.0 & 23/100 & 38/100\\
\midrule
ReCo (Ours) & RL+MPC & \textbf{0.135 $\pm$ 0.091} & \textbf{23.8 $\pm$ 13.7} & \textbf{94.0} & \textbf{75/100} & \textbf{3/100}\\
\bottomrule
\end{tabular}
\end{table*}

\begin{figure*}[!t]
\centering
\includegraphics[width=0.85\textwidth]{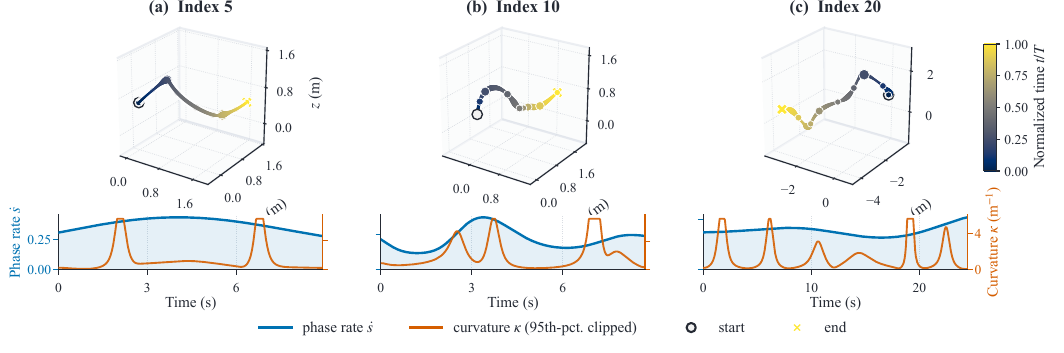}
\caption{Representative benchmark trajectories. Top: 3D EE paths colored by normalized time. Bottom: reference phase rate $\dot{s}$ and path curvature $\kappa$ (clipped at the 95th percentile for display). Spatial extent, curvature, and traversal timing vary across trajectories.}
\label{fig:trajectory-library}
\end{figure*}

Figure~\ref{fig:trajectory-library} illustrates the spatial and temporal demands of the benchmark. ReCo achieves the lowest mean position and orientation RMSE in Table~\ref{tab:formal-benchmark}, 28.7\% and 27.4\% below the best baseline on each metric (Pure MPC and Ma et al., respectively). Paired Wilcoxon signed-rank tests over the 100 runs favor ReCo against every baseline on both metrics ($p<10^{-3}$ after Holm correction). Decomposed into lateral-path and arc-length timing components, the mean position errors of ReCo are 0.1261 and 0.1701~m, compared with 0.2270 and 0.3608~m for Ma et al.

ReCo also completes more of each reference: its mean progress is 11.4 percentage points above the next-best controller, at a mean base speed (0.205~m/s) comparable to Ma et al. (0.217~m/s) and above Pure MPC (0.158~m/s). It succeeds in 75/100 runs (39/50 nominal and 36/50 push), compared with 42/100 for RoboDuet+PP, the strongest baseline by success count, and records the fewest falls. Thus, the tracking improvement accompanies greater completion and fewer falls.

\textit{Baseline behavior.} RoboDuet+PP and Visual WBC+PP complete at most 65.7\% of the reference on average, with position RMSE 5.0 and 10.8 times that of ReCo. Their externally supplied base paths are not jointly optimized with arm motion for the imposed EE reference. These results are consistent with a benefit from planning base displacement and arm motion together, although the comparison does not isolate planning from differences in the underlying controllers.

UMI on Legs is trained to track task-frame trajectories drawn from manipulation demonstrations~\cite{ha2024umilegs}. On in-distribution trajectories, our retrained policy achieves position and orientation RMSEs of 0.035~m and $4.7^\circ$, respectively. On the benchmark trajectories, it falls in 80/100 runs. This contrast highlights the challenge of generalizing from demonstration-derived motions to the diverse geometry and timing of independently generated references.

\textit{Disturbance response.} ReCo's mean position RMSE increases from 0.1326~m in nominal runs to 0.1371~m under the applied push (3.4\%), compared with a 6.4\% increase for Ma et al. ReCo records two falls in nominal runs and one in push runs. Aggregate tracking error thus changes little under the tested impulse, which covers one direction, duration, and application schedule.

\begin{figure}[!t]
\centering
\includegraphics[width=0.85\columnwidth]{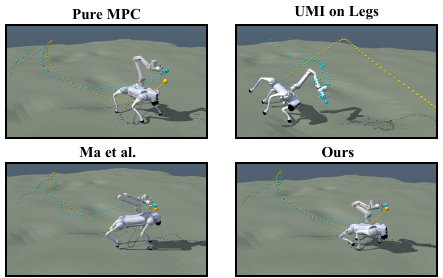}
\par{\small (a) Recorded simulation states\par}\smallskip
\includegraphics[width=\columnwidth]{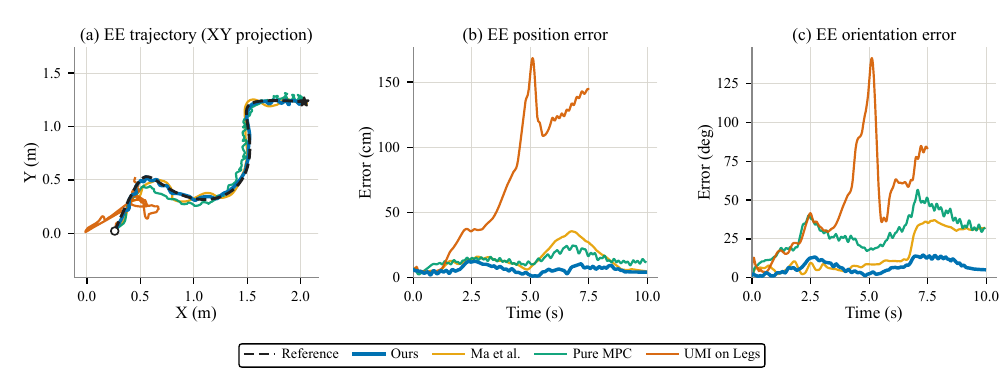}
\par{\small (b) EE paths and tracking errors\par}
\caption{Selected controllers on one nominal benchmark trajectory. (a) Final recorded states, which occur at different times after early termination; yellow and cyan mark reference and actual EE paths. (b) Equal-scale XY paths and tracking errors (orientation in radians); curves use a five-sample moving average; RMSE uses unfiltered samples. UMI on Legs terminates after a fall.}
\label{fig:mujoco-comparison}
\end{figure}

Figure~\ref{fig:mujoco-comparison} shows a selected 3.43~m trajectory with an 8.92~s reference duration. ReCo achieves 0.0695~m position and $8.20^\circ$ orientation RMSE, less than half the error of either Pure MPC or Ma et al. on both metrics. Both ReCo and Ma et al. succeed. Pure MPC reaches the endpoint without satisfying the hold requirement; UMI on Legs falls at 12.2\% progress.

\subsection{Response Shaping and Model Identification}
\label{sec:ablation}
\begin{table}[t]
\centering
\caption{Factorial ablation over response shaping and model identification. Without identification, MPC assumes instantaneous command execution. All metrics are lower-is-better; the best entry per row is bold.}
\label{tab:shaping-identification}
\small
\setlength{\tabcolsep}{3pt}
\begin{tabular}{lrrrr}
\toprule
Response shaping & -- & -- & $\checkmark$ & $\checkmark$\\
Model identification & -- & $\checkmark$ & -- & $\checkmark$\\
\midrule
Position RMSE (m) & 0.257 & 0.232 & 0.166 & \textbf{0.135}\\
Orientation RMSE ($^\circ$) & 21.7 & 24.3 & \textbf{21.5} & 23.8\\
EE jerk (m/s$^3$) & 252.1 & 250.1 & 261.0 & \textbf{249.7}\\
Base jerk (m/s$^3$) & 145.6 & \textbf{140.3} & 155.4 & 149.9\\
Leg torque RMS (N$\cdot$m) & 6.41 & 6.40 & 6.46 & \textbf{6.30}\\
\bottomrule
\end{tabular}
\end{table}

Table~\ref{tab:shaping-identification} reports a $2\times2$ factorial ablation of response shaping and model identification on the same benchmark. We refer to policies trained without and with response shaping as \emph{unshaped} and \emph{shaped}, respectively. Without identification, MPC assumes instantaneous, exact command execution.

Both response shaping and model identification reduce position error, and their combination achieves the lowest error. Shaping alone reduces position RMSE from 0.257 to 0.166~m (35.4\%). For the shaped policy, identification further reduces the error to 0.135~m (18.7\%), compared with a 9.7\% reduction for the unshaped policy. The prediction experiments in Sec.~\ref{sec:prediction} examine how shaping changes model accuracy.

Model identification improves position accuracy but increases orientation RMSE from $21.7^\circ$ to $24.3^\circ$ for the unshaped policy and from $21.5^\circ$ to $23.8^\circ$ for the shaped policy. This trade-off may reflect attitude dynamics and translation--rotation coupling omitted by the per-channel response model. All four configurations have lower mean orientation error than the best baseline in Table~\ref{tab:formal-benchmark}. ReCo has the lowest EE jerk and leg torque RMS among the ablations, but not the lowest base jerk. Identification therefore primarily improves position accuracy in these runs.

\subsection{Response Prediction and Repeatability}
\label{sec:prediction}
We compare the unshaped and shaped policies on 36 command trajectories held out from identification. A response model with the same structure is fitted independently to each frozen policy's response. Predictions start from measured states and are propagated under recorded commands without intermediate measurement correction. We evaluate forward, lateral, and yaw-velocity errors at 0.25, 0.50, and 1.00~s. MSE is normalized by each channel's variance and averaged equally across trajectories, channels, and horizons.

\begin{figure}[!t]
\centering
\includegraphics[width=\columnwidth]{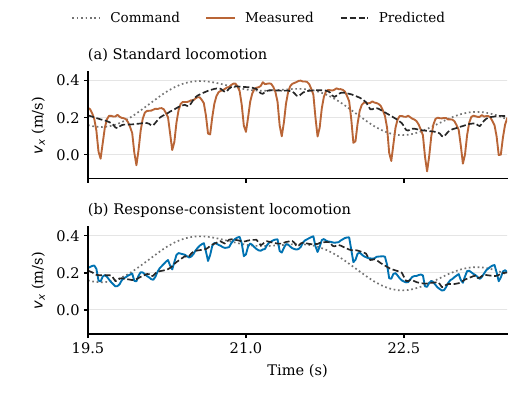}
\caption{Forward-velocity prediction for the unshaped and shaped policies, labeled standard and response-consistent in the legend. Measured responses and fixed 0.50~s forecasts are plotted at forecast target times on identical axes. Table~\ref{tab:response-summaries} reports aggregate errors.}
\label{fig:predictability}
\end{figure}

Shaping reduces normalized prediction MSE from 0.105 to 0.042 (59.6\%). Table~\ref{tab:response-summaries} shows improvement at all three horizons, with little change between 0.50 and 1.00~s. Figure~\ref{fig:predictability} illustrates the command transient and corresponding forecasts.

\begin{table}[t]
\centering\small
\caption{Prediction accuracy and response repeatability. Response shaping lowers prediction error at every horizon and reduces the sensitivity of the forward-velocity response to arm motion.}
\label{tab:response-summaries}
\renewcommand{\arraystretch}{1.12}
\begin{tabular*}{\columnwidth}{@{\extracolsep{\fill}}lcc@{}}
\toprule
Condition & Unshaped & Shaped\\
\midrule
\multicolumn{3}{@{}l}{\textit{Normalized prediction MSE}}\\[2pt]
Horizon 0.25~s & 0.0958 & \textbf{0.0395}\\
Horizon 0.50~s & 0.1091 & \textbf{0.0442}\\
Horizon 1.00~s & 0.1097 & \textbf{0.0434}\\
\addlinespace[5pt]
\multicolumn{3}{@{}l}{\textit{Forward-velocity residual RMS (m/s)}}\\[2pt]
Arm held & 0.0135 & \textbf{0.0099}\\
Arm moving & 0.0345 & \textbf{0.0139}\\
Moving $-$ held & 0.0210 & \textbf{0.0040}\\
\bottomrule
\end{tabular*}
\par\smallskip
\begin{minipage}{\columnwidth}\footnotesize Prediction: 36 command trajectories held out from identification, three velocity channels. Residual: means over 12 matched conditions after gait-phase conditioning.\end{minipage}
\end{table}

\begin{figure}[!t]
\centering
\includegraphics[width=\columnwidth]{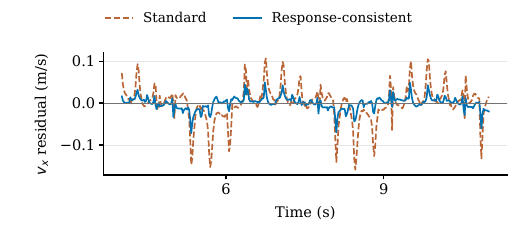}
\caption{Phase-conditioned forward-velocity residual in a selected moving-arm run. Table~\ref{tab:response-summaries} reports mean residual RMS across the matched conditions.}
\label{fig:arm-repeatability}
\end{figure}

We assess repeatability under 12 matched speed and initial-phase conditions, with the arm held stationary or executing periodic six-joint motion. We compute residual RMS after subtracting phase-conditioned means over $[4,11)$~s using 16 gait-phase bins. Under arm motion, forward-velocity residual RMS decreases from 0.0345 to 0.0139~m/s with shaping (Fig.~\ref{fig:arm-repeatability}). Moving the arm increases residual RMS by 0.0040~m/s for the shaped policy, compared with 0.0210~m/s for the unshaped policy. These results indicate reduced sensitivity of forward response to the tested arm motions without an explicit wrench model. Height residuals do not show the same improvement, despite height being included in the cross-domain objective.

\subsection{Gait-Periodic Prediction}
\label{sec:phase-prediction}
\begin{figure}[!t]
\centering
\includegraphics[width=\columnwidth]{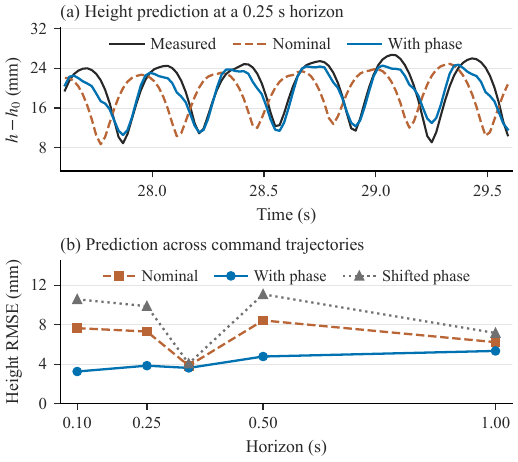}
\caption{Gait-periodic height prediction. (a) Measured height relative to the nominal offset and fixed 0.25~s forecasts. (b) Height RMSE across 36 trajectories, including a shifted-phase control. The reported aggregate uses 0.10, 0.25, and 0.34~s; 0.50 and 1.00~s are also shown. All forecasts use causal phase extrapolation.}
\label{fig:phase-prediction}
\end{figure}

With the policy and the non-periodic part of the response model fixed, we evaluate the gait-periodic component in~\eqref{eq:gait}. Phase is extrapolated from information available before each forecast starts, and an initial offset matches the measured state. Across 36 trajectories and horizons of 0.10, 0.25, and 0.34~s, height RMSE decreases from 6.49 to 3.58~mm (44.8\%). Squared errors are averaged across trajectories and horizons before taking the square root. Figure~\ref{fig:phase-prediction} shows closer alignment with the measured gait cycle; a deliberately shifted phase degrades prediction, supporting the importance of phase alignment.

In the closed-loop component comparison, adding the gait-periodic component reduces position RMSE by 5.5\%, orientation RMSE by 6.9\%, and vertical tracking-error RMS in the 1.5--8~Hz trot-excitation band by 6.5\%.

\subsection{Hardware Experiments}
\label{sec:hardware}
The Go2--X5 platform integrates a Unitree Go2, an ARX X5 six-DoF arm, and a Jetson Orin NX onboard computer. The control stack uses ROS~2 Foxy~\cite{macenski2022ros2} and runs entirely onboard. The complete software stack, including concurrent programs, uses an average of 75\% CPU and 43\% RAM.

\begin{figure}[!t]
\centering
\includegraphics[width=0.9\columnwidth]{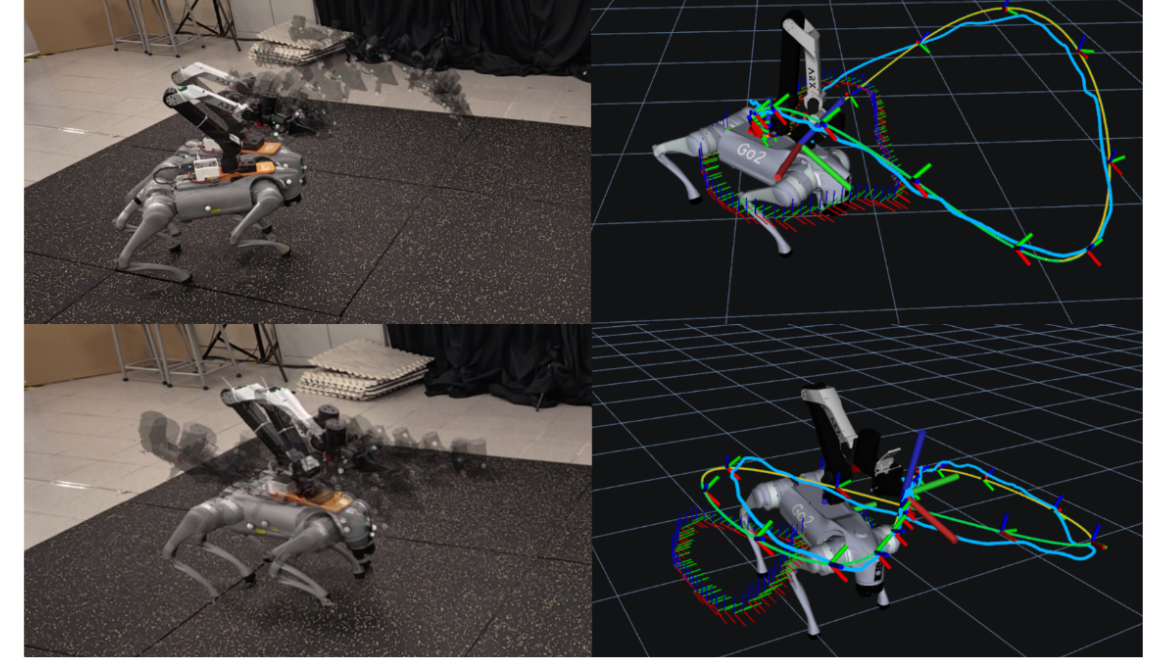}
\caption{Representative trajectories from the motion-capture experiments. Left: hardware execution. Right: reference and measured EE paths.}
\label{fig:hardware-experiments}
\end{figure}

\textit{Motion-capture tracking.} To bound the hardware accuracy without state-estimation drift, motion capture supplies base-state feedback and EE ground truth on eight trajectories (Fig.~\ref{fig:hardware-experiments}). Mean position and orientation RMSEs are 0.0760~m and $9.93^\circ$; the shaped policy lowers them by 29.1\% and 7.4\% relative to the unshaped policy.

Outside the motion-capture arena, world-frame EE trajectories for object organization, cabinet-door opening, and pick-and-place are prerecorded using VR teleoperation (Fig.~\ref{fig:teaser}). ReCo completes these tasks during locomotion, with mean position and orientation RMSEs of 0.153~m and $24.9^\circ$ over three runs per task. EE poses are reconstructed from the onboard odometry and arm forward kinematics, so these metrics characterize tracking in the estimated world frame, without an independent measure of odometry drift. The experiments demonstrate onboard execution of ReCo on physical manipulation tasks.

\section{Conclusion}
We presented ReCo, a framework that couples response-consistent locomotion with policy-aware MPC for legged manipulation. Response shaping and an identified closed-loop response model allowed MPC to jointly plan locomotion commands and arm motion. On the simulation benchmark, ReCo reduced position and orientation RMSE by 28.7\% and 27.4\% relative to the best baseline for each metric and succeeded in 75 of 100 runs, compared with 42 for the strongest baseline. Response shaping lowered normalized prediction MSE by 59.6\%, and hardware experiments demonstrated onboard manipulation during locomotion. Model identification reduced position error but increased orientation error, and shaping did not improve height repeatability. Future work could address both by modeling coupled arm--base dynamics in the response model and by shaping the height response, and could extend the predictive interface to perceptive locomotion.

\bibliographystyle{IEEEtran}
\bibliography{references}

@inproceedings{ha2024umilegs,
  author = {Huy Ha and Yihuai Gao and Zipeng Fu and Jie Tan and Shuran Song},
  title = {{UMI-on-Legs}: Making Manipulation Policies Mobile with Manipulation-Centric Whole-body Controllers},
  booktitle = {Proceedings of the 8th Conference on Robot Learning},
  series = {Proceedings of Machine Learning Research}, volume = {270},
  pages = {5254--5270}, year = {2025}
}

@inproceedings{liu2024visual,
  author = {Minghuan Liu and Zixuan Chen and Xuxin Cheng and Yandong Ji and Ri-Zhao Qiu and Ruihan Yang and Xiaolong Wang},
  title = {Visual Whole-Body Control for Legged Loco-Manipulation},
  booktitle = {Proceedings of the 8th Conference on Robot Learning},
  series = {Proceedings of Machine Learning Research}, volume = {270},
  pages = {234--257}, year = {2025}
}

@article{miki2022perceptive,
  author = {Takahiro Miki and Joonho Lee and Jemin Hwangbo and Lorenz Wellhausen and Vladlen Koltun and Marco Hutter},
  title = {Learning Robust Perceptive Locomotion for Quadrupedal Robots in the Wild},
  journal = {Science Robotics}, volume = {7}, number = {62}, pages = {eabk2822}, year = {2022},
  doi = {10.1126/scirobotics.abk2822}
}

@inproceedings{margolis2023wtw,
  author = {Gabriel B. Margolis and Pulkit Agrawal},
  title = {{Walk These Ways}: Tuning Robot Control for Generalization with Multiplicity of Behavior},
  booktitle = {Proceedings of the 6th Conference on Robot Learning},
  series = {Proceedings of Machine Learning Research}, volume = {205}, pages = {22--31}, year = {2023}
}

@inproceedings{bellicoso2019alma,
  author = {C. Dario Bellicoso and Koen Kr{\"a}mer and Markus St{\"a}uble and Dhionis Sako and Fabian Jenelten and Marko Bjelonic and Marco Hutter},
  title = {{ALMA}---Articulated Locomotion and Manipulation for a Torque-Controllable Robot},
  booktitle = {IEEE International Conference on Robotics and Automation},
  pages = {8477--8483}, year = {2019}, doi = {10.1109/ICRA.2019.8794273}
}

@article{pankert2020perceptive,
  author = {Johannes Pankert and Marco Hutter},
  title = {Perceptive Model Predictive Control for Continuous Mobile Manipulation},
  journal = {IEEE Robotics and Automation Letters}, volume = {5}, number = {4}, pages = {6177--6184},
  year = {2020}, doi = {10.1109/LRA.2020.3010721}
}

@article{grandia2022perceptive,
  author = {Ruben Grandia and Fabian Jenelten and Shaohui Yang and Farbod Farshidian and Marco Hutter},
  title = {Perceptive Locomotion Through Nonlinear Model-Predictive Control},
  journal = {IEEE Transactions on Robotics}, volume = {39}, number = {5}, pages = {3402--3421},
  year = {2023}, doi = {10.1109/TRO.2023.3275384}
}

@inproceedings{fu2023deep,
  author = {Zipeng Fu and Xuxin Cheng and Deepak Pathak},
  title = {Deep Whole-Body Control: Learning a Unified Policy for Manipulation and Locomotion},
  booktitle = {Proceedings of the 6th Conference on Robot Learning},
  series = {Proceedings of Machine Learning Research}, volume = {205}, pages = {138--149}, year = {2023}
}

@article{pan2024roboduet,
  author = {Guoping Pan and Qingwei Ben and Zhecheng Yuan and Guangqi Jiang and Yandong Ji and Shoujie Li and Jiangmiao Pang and Houde Liu and Huazhe Xu},
  title = {{RoboDuet}: Learning a Cooperative Policy for Whole-Body Legged Loco-Manipulation},
  journal = {IEEE Robotics and Automation Letters}, volume = {10}, number = {5}, pages = {4564--4571},
  year = {2025}, doi = {10.1109/LRA.2025.3551230}
}

@article{ma2022combining,
  author = {Yuntao Ma and Farbod Farshidian and Takahiro Miki and Joonho Lee and Marco Hutter},
  title = {Combining Learning-Based Locomotion Policy with Model-Based Manipulation for Legged Mobile Manipulators},
  journal = {IEEE Robotics and Automation Letters}, volume = {7}, number = {2},
  pages = {2377--2384}, year = {2022}, doi = {10.1109/LRA.2022.3143567}
}

@article{cheng2025rambo,
  author = {Jin Cheng and Dongho Kang and Gabriele Fadini and Guanya Shi and Stelian Coros},
  title = {{RAMBO}: {RL}-Augmented Model-Based Whole-Body Control for Loco-Manipulation},
  journal = {IEEE Robotics and Automation Letters}, volume = {10}, number = {9}, pages = {9462--9469},
  year = {2025}, doi = {10.1109/LRA.2025.3594984}
}

@misc{kim2026koopman,
  author = {Jeonghwan Kim and Yunhai Han and Harish Ravichandar and Sehoon Ha},
  title = {Safe Navigation of Bipedal Robots via {Koopman} Operator-Based Model Predictive Control},
  year = {2026}, howpublished = {arXiv:2409.14736v3},
  note = {Version 3; first submitted in 2024}
}

@inproceedings{dicarlo2018cheetah,
  author = {Jared {Di Carlo} and Patrick M. Wensing and Benjamin Katz and Gerardo Bledt and Sangbae Kim},
  title = {Dynamic Locomotion in the {MIT Cheetah 3} Through Convex Model-Predictive Control},
  booktitle = {IEEE/RSJ International Conference on Intelligent Robots and Systems},
  pages = {1--9}, year = {2018}, doi = {10.1109/IROS.2018.8594448}
}

@inproceedings{tan2018sim,
  author = {Jie Tan and Tingnan Zhang and Erwin Coumans and Atil Iscen and Yunfei Bai and Danijar Hafner and Steven Bohez and Vincent Vanhoucke},
  title = {Sim-to-Real: Learning Agile Locomotion for Quadruped Robots},
  booktitle = {Proceedings of Robotics: Science and Systems},
  year = {2018}, doi = {10.15607/RSS.2018.XIV.010}
}

@article{hwangbo2019agile,
  author = {Jemin Hwangbo and Joonho Lee and Alexey Dosovitskiy and Dario Bellicoso and Vassilios Tsounis and Vladlen Koltun and Marco Hutter},
  title = {Learning Agile and Dynamic Motor Skills for Legged Robots},
  journal = {Science Robotics}, volume = {4}, number = {26}, pages = {eaau5872},
  year = {2019}, doi = {10.1126/scirobotics.aau5872}
}

@article{lee2020terrain,
  author = {Joonho Lee and Jemin Hwangbo and Lorenz Wellhausen and Vladlen Koltun and Marco Hutter},
  title = {Learning Quadrupedal Locomotion over Challenging Terrain},
  journal = {Science Robotics}, volume = {5}, number = {47}, pages = {eabc5986},
  year = {2020}, doi = {10.1126/scirobotics.abc5986}
}

@inproceedings{kumar2021rma,
  author = {Ashish Kumar and Zipeng Fu and Deepak Pathak and Jitendra Malik},
  title = {{RMA}: Rapid Motor Adaptation for Legged Robots},
  booktitle = {Proceedings of Robotics: Science and Systems},
  year = {2021}, doi = {10.15607/RSS.2021.XVII.011}
}

@inproceedings{rudin2022minutes,
  author = {Nikita Rudin and David Hoeller and Philipp Reist and Marco Hutter},
  title = {Learning to Walk in Minutes Using Massively Parallel Deep Reinforcement Learning},
  booktitle = {Proceedings of the 5th Conference on Robot Learning},
  series = {Proceedings of Machine Learning Research}, volume = {164},
  pages = {91--100}, year = {2022}
}

@inproceedings{portela2025pose,
  author = {Tifanny Portela and Andrei Cramariuc and Mayank Mittal and Marco Hutter},
  title = {Whole-Body End-Effector Pose Tracking},
  booktitle = {IEEE International Conference on Robotics and Automation},
  pages = {11205--11211}, year = {2025}, doi = {10.1109/ICRA55743.2025.11127964}
}

@inproceedings{portela2024force,
  author = {Tifanny Portela and Gabriel B. Margolis and Yandong Ji and Pulkit Agrawal},
  title = {Learning Force Control for Legged Manipulation},
  booktitle = {IEEE International Conference on Robotics and Automation},
  year = {2024}, doi = {10.1109/ICRA57147.2024.10611066}
}

@article{hewing2020learning,
  author = {Lukas Hewing and Kim P. Wabersich and Marcel Menner and Melanie N. Zeilinger},
  title = {Learning-Based Model Predictive Control: Toward Safe Learning in Control},
  journal = {Annual Review of Control, Robotics, and Autonomous Systems},
  volume = {3}, pages = {269--296}, year = {2020}, doi = {10.1146/annurev-control-090419-075625}
}

@misc{qmcontrol,
  author = {{skywoodsz}},
  title = {{qm\_control}: Quadruped Manipulator Controller Using Model Predictive Control and Whole-Body Control},
  howpublished = {Code: \url{https://github.com/skywoodsz/qm_control}},
  note = {Accessed: September 16, 2026}
}

@misc{schulman2017ppo,
  author = {John Schulman and Filip Wolski and Prafulla Dhariwal and Alec Radford and Oleg Klimov},
  title = {Proximal Policy Optimization Algorithms}, year = {2017},
  howpublished = {arXiv:1707.06347}
}

@article{macenski2022ros2,
  author = {Steven Macenski and Tully Foote and Brian Gerkey and Chris Lalancette and William Woodall},
  title = {Robot Operating System 2: Design, Architecture, and Uses in the Wild},
  journal = {Science Robotics}, volume = {7}, number = {66}, pages = {eabm6074},
  year = {2022}, doi = {10.1126/scirobotics.abm6074}
}

@inproceedings{todorov2012mujoco,
  author = {Todorov, Emanuel and Erez, Tom and Tassa, Yuval},
  title = {{MuJoCo}: A physics engine for model-based control},
  booktitle = {IEEE/RSJ International Conference on Intelligent Robots and Systems},
  year = {2012}, pages = {5026--5033}
}

@inproceedings{makoviychuk2021isaac,
  author = {Makoviychuk, Viktor and Wawrzyniak, Lukasz and Guo, Yunrong and Lu, Michelle and Storey, Kier and Macklin, Miles and Hoeller, David and Rudin, Nikita and Allshire, Arthur and Handa, Ankur and State, Gavriel},
  title = {{Isaac Gym}: High performance {GPU}-based physics simulation for robot learning},
  booktitle = {Proceedings of the Neural Information Processing Systems Track on Datasets and Benchmarks},
  year = {2021}
}

@misc{farshidian2017ocs2,
  author = {Farshidian, Farbod and others},
  title = {{OCS2}: An open source library for optimal control of switched systems},
  howpublished = {[Online]. Available: \url{https://github.com/leggedrobotics/ocs2}}
}
\end{document}